\documentclass[10pt,twocolumn,letterpaper]{article}

\usepackage{cvpr}              

\usepackage[dvipsnames]{xcolor}

\definecolor{cvprblue}{rgb}{0.21,0.49,0.74}
\usepackage[pagebackref,breaklinks,colorlinks,citecolor=cvprblue]{hyperref}
\usepackage{adjustbox}

\usepackage{multirow,algorithm}
\usepackage{adjustbox}
\usepackage{algpseudocode}

\def\paperID{6178} 
\def\confName{CVPR}
\def\confYear{2024}

\title{Closed-Loop Unsupervised Representation Disentanglement with \\ $\beta$-VAE Distillation and Diffusion Probabilistic Feedback}

\author{
Xin Jin\textsuperscript{\rm 1, *, †},
    Bohan Li\textsuperscript{\rm 1, 2, *},
BAAO Xie\textsuperscript{\rm 1},\\
Wenyao Zhang\textsuperscript{\rm 1, 2},
    Jinming Liu\textsuperscript{\rm 1, 2},
Ziqiang Li\textsuperscript{\rm 1, 2},
Tao Yang\textsuperscript{\rm 3}, 
    Wenjun Zeng\textsuperscript{\rm 1}, \\
    \textsuperscript{\rm 1}Ningbo Institute of Digital Twin, Eastern Institute of Technology, Ningbo, China \\
    \textsuperscript{\rm 2}Shanghai Jiao Tong University, Shanghai, China, 
    \textsuperscript{\rm 3}Xi'an Jiaotong University, Xi'an, China \\
    \textsuperscript{*}These authors contributed equally to this work \\
    \textsuperscript{†}Corresponding: jinxin@eitech.edu.cn \\
}

\begin{document}
\maketitle

\newcommand{\vect}[1]{\ensuremath{\mathbf{#1}}}
\newcommand{\norm}[1]{\left\lVert#1\right\rVert}
\newcommand{\zsem}{\vect{z}_\text{sem}}
\newcommand{\xT}{\vect{x}_T}
\newcommand{\xt}{\vect{x}_t}
\newcommand{\xzero}{\vect{x}_0}
\newcommand{\xtone}{\vect{x}_{t-1}}
\newcommand{\Ndist}{\mathcal{N}(\vect{0}, \mathbf{I})}

\newcommand{\bohan}{\textcolor{red}}
\newcommand{\baao}{\textcolor{blue}}
\newcommand{\wenyao}{\textcolor{green}}
\newcommand{\jinming}{\textcolor{yellow}}
\newcommand{\ziqiang}{\textcolor{pink}}







\begin{abstract}

Representation disentanglement may help AI fundamentally understand the real world and thus benefit both discrimination and generation tasks. It currently has at least three unresolved core issues: (i) heavy reliance on label annotation and synthetic data --- causing poor generalization on natural scenarios; (ii) heuristic/hand-craft disentangling constraints make it hard to adaptively achieve an optimal training trade-off; (iii) lacking reasonable evaluation metric, especially for the real label-free data. To address these challenges, we propose a \textbf{C}losed-\textbf{L}oop unsupervised representation \textbf{Dis}entanglement approach dubbed \textbf{CL-Dis}. Specifically, we use diffusion-based autoencoder (Diff-AE) as a backbone while resorting to $\beta$-VAE as a co-pilot to extract semantically disentangled representations. The strong generation ability of diffusion model and the good disentanglement ability of VAE model are complementary. To strengthen disentangling, VAE-latent distillation and diffusion-wise feedback are interconnected in a closed-loop system for a further mutual promotion. Then, a self-supervised \textbf{Navigation} strategy is introduced to identify interpretable semantic directions in
the disentangled latent space. Finally, a new metric based on content tracking is designed to evaluate the disentanglement effect. Experiments demonstrate the superiority of CL-Dis on applications like real image manipulation and visual analysis.

\end{abstract}
\vspace{-5mm}
\section{Introduction}
\label{sec:intro}

Disentangled representation learning (DRL)~\cite{higgins2018towards} learns the underlying explainable factors behind the observed data, where the different latent factors correspond to different properties, respectively. This matches the way humans perceive this world that we understand an object from its various properties (e.g., shape, size, color, etc.)~\cite{egan1989memory,fleming2014visual,yan2016attribute2image,pearson2019human,bao2022generative}. Thus, DRL is thought to be one of the possible ways for AI to understand the world fundamentally, thus achieving Artificial General Intelligence (AGI)~\cite{bengio2013representation,lake2017building,wang2022disentangled}. 



\begin{figure}
    \centering
    \includegraphics[width=\linewidth]{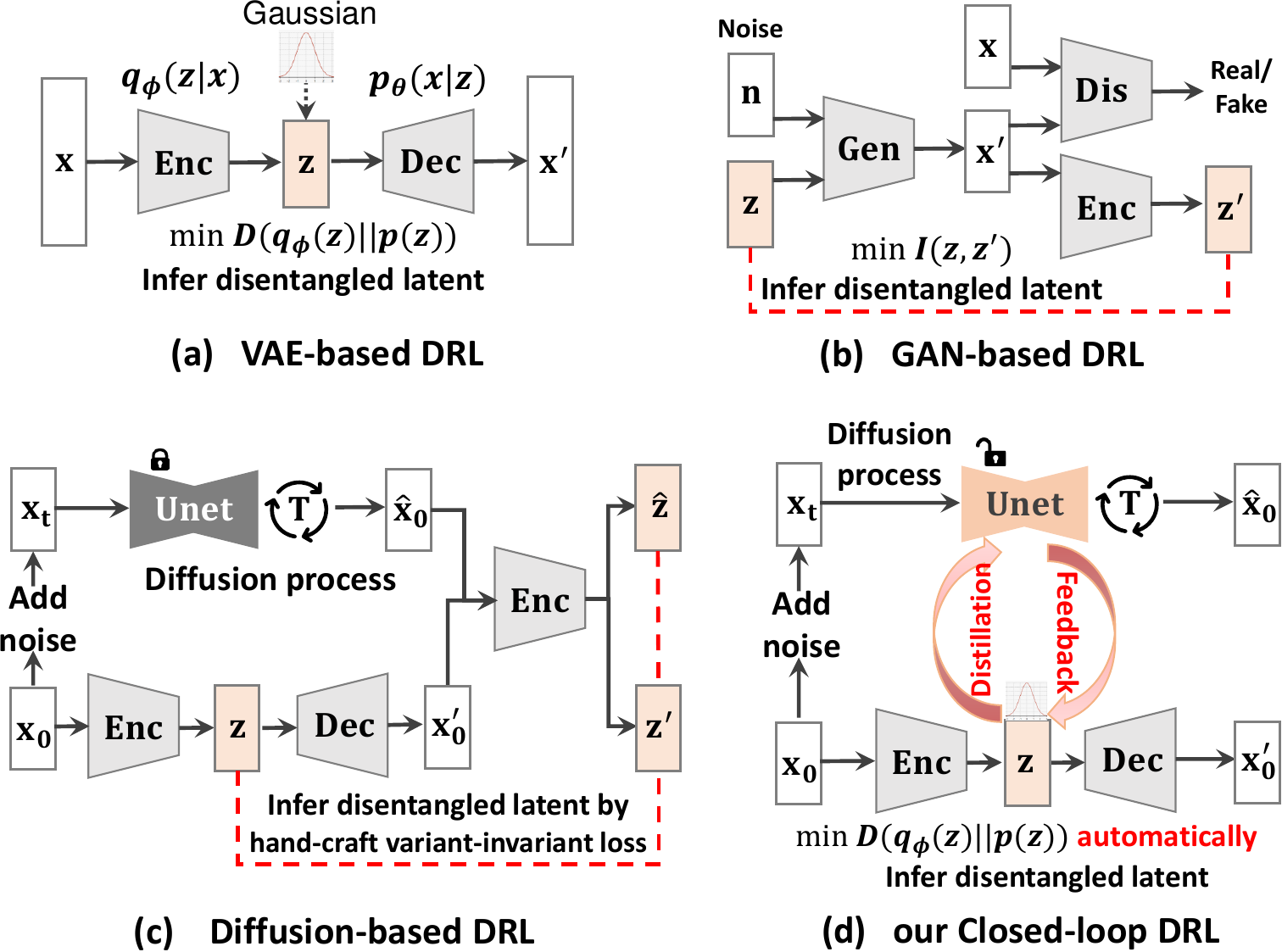}
    \vspace{-6mm}
    \caption{The illustration of disentangled representation learning (DRL) frameworks: (a) VAE-based, (b) GAN-based, (c) Diffusion-based, and (d) our proposed Closed-loop approach. We can intuitively see that previous works all relied on heuristic hand-craft loss constraints to infer disentangled latent, only our method leverages two kinds of generative branches to build a cycle system for mutually-promoting automatically disentangled learning.}
    \label{fig:motivation}
    \vspace{-5mm}
\end{figure}

Most existing DRL methods learn the disentangled representation based on generative models, such as VAE~\cite{kingma2013auto,higgins2016beta,burgess2018understanding,kumar2017variational,kim2018disentangling,chen2018isolating,dupont2018learning,kim2019relevance,shao2020controlvae}, GAN~\cite{goodfellow2014generative,chen2016infogan,tran2017disentangled,lin2019infogancr,jeon2021ibGAN,zhu2021and}, and even recently popular diffusion~\cite{kwon2023diffusion,wu2023uncovering,yang2023disdiff,wu2023not,xu2023versatile}. Among them, the VAE-based DRL methods always encourage the disentanglement of the latent variables by forcing the variational posterior to be closer to the factorized
Gaussian prior (see Figure~\ref{fig:motivation}(a)), but they generally need to balance a hard inherent trade-off between the disentangling ability and generating quality~\cite{higgins2016beta,chen2018isolating,kim2018disentangling}. As shown in Figure~\ref{fig:motivation}(b), the GAN-based DRL methods tend to use the mutual information as disentanglement constraints, but they typically suffer from the problem of reconstruction quality due to the gan-inversion difficulty~\cite{wang2022high,yang2023disdiff}. Besides, VAE- and GAN-based generative models are prone to learn the factorized or conditioned statistics bias~\cite{ho2022classifier}, which essentially conflicts with the ultimate purpose of a generalizable disentanglement that makes AI understand the world like humans.




Moreover, given the recently popular diffusion probabilistic models (DPMs) show superiority in image generation quality~\cite{ho2020denoising,song2020denoising} and unsupervised controlling (i.e., learning meaningful representations) of images~\cite{preechakul2022diffusion}, researchers have started to investigate the possibility of DRL in DPMs. Asyrp~\cite{kwon2023diffusion} proposes an asymmetric reverse process to discover the semantic latent space in frozen pre-trained diffusion models, which reveals that DPMs already have a semantic latent space and thus more suitable for learning disentangled semantics. The co-current work of DisDiff~\cite{yang2023disdiff} shown in Figure~\ref{fig:motivation}(c) uses the gradient field of frozen DPMs to achieve a vector-wise disentangled representation learning. However, this work applies an external heuristic hand-craft variant-invariant loss~\cite{locatello2019challenging,yang2023disdiff} to enforce the disentanglement processes, resulting in reduced model flexibility and posing challenges for optimization.




In this paper, we propose to leverage two kinds of generative branches of the diffusion model and VAE model, to build a cycle system for mutually-promoting DRL. As shown in Figure~\ref{fig:motivation}(d)), we construct an unsupervised \textbf{C}losed-\textbf{L}oop \textbf{Dis}entanglement framework dubbed \textbf{CL-Dis}, which is built upon a diffusion-based autoencoder (Diff-AE) backbone while combining with a co-pilot of $\beta$-VAE, to together learn the meaningful disentangled
representations. To collaborate two branches well and facilitate feature disentangling, we encourage Diff-AE and $\beta$-VAE to \textbf{complement each other, achieving a win-win effect:} we use the pre-trained latent of VAE to guide a semantic-aware reverse diffusion process with a distillation loss. Then, the gradually increasing information capability during the diffusion process is taken as feedback, in turn, to progressively enforce the inner latent of VAE more disentangling. Next, a self-supervised Navigation strategy is introduced to clearly identify each factor's semantic meanings, where shifting along a disentangled semantic direction will result in \textit{continuously changed generations}. Finally, we further design a label-free metric based on \textbf{``changes tracking''} to quantitatively measure the disentanglement of learned features, wherein we use optical flow to reflect image variation degree objectively. Our main contributions are summarized:

\begin{itemize}

    \item We build a new unsupervised representation disentanglement framework CL-Dis, which has a mutually-promoting closed-loop architecture driven by diffusion and VAE models. They complement each other with a distillation loss and a new feedback loss, and thus empower a more controllable and stronger DRL capability.

    \item We introduce a self-supervised method in CL-Dis to clearly identify each factor's semantic meanings by navigating directions in the learned disentangled latent space, leading to coherent generated variations. This makes the generation more explainable and fine-controllable.

    \item We design a new label-free metric to quantitatively measure the interpretability and disentanglement of learned features, and provide an example of immediate practical benefit from our work. Namely, we experimentally show how to exploit our CL-Dis for generative manipulation of real images and visual discrimination tasks.
    
\end{itemize}

\vspace{-3mm}
\section{Related Works}
\label{sec:related}
\vspace{-1mm}

\subsection{Disentangled Representation Learning (DRL)} 

The concept of DRL was introduced by Bengio et al.~\cite{bengio2013representation} in 2013 and was believed helpful in different tasks in practical applications like image generation~\cite{kingma2013auto,chen2018isolating,kumar2017variational,kim2018disentangling,kim2019relevance,wu2021stylespace,chen2016infogan,lin2019infogancr}, image editing/translation~\cite{gonzalez2018image,lee2018diverse,liu2021smoothing,wang2023stylediffusion}, NLP~\cite{wu2020improving,he2017unsupervised,cheng2020improving} and multimodal applications~\cite{hsu2018disentangling,tsai2018learning,zhang2022learning,xu2022predict,materzynska2022disentangling}. 
{Different from image generation or editing, the core of this task is to fully understand the latent factors of the model to enhance its fine-grained controllability~\cite{chen2018isolating,kumar2017variational,kim2018disentangling,kim2019relevance,wu2021stylespace}.}

Most VAE-based DRL approaches like $\beta$-VAE~\cite{higgins2016beta}, 
DIP-VAE~\cite{kumar2017variational}, FactorVAE~\cite{kim2018disentangling}, etc., achieve unsupervised disentanglement by the direct constraints on probabilistic distributions~\cite{chen2018isolating,kim2018disentangling,higgins2016beta}. While straightforward, Locatello et al.~\cite{locatello2019challenging} point out that is not enough and emphasize the need for extra inductive bias. Similarly, Burgess~\cite{burgess2018understanding} proposes to progressively
increase the information capacity of the latent code in $\beta$-VAE. Yang et al.~\cite{yang2021towards} use symmetry properties modeled by group theory as inductive bias. Except for VAE, there are also some DRL works based on GAN, e.g., leveraging mutual information~\cite{chen2016infogan}, self-supervised contrastive regularizer~\cite{lin2019infogancr}, spatial constriction~\cite{zhu2021and}, and combining with other pre-trained generative models~\cite{lee2020high,ren2021learning}.

Fast-developing generative diffusion probabilistic models (DPMs) have accelerated the DRL exploration. LCG-DM~\cite{kim2022unsupervised} uses a vanilla VAE in DPMs to extract semantics codes for controllable generation, but ignores the interaction between these two structures. Asyrp~\cite{kwon2023diffusion} discovers the semantic latent space in frozen pre-trained DPMs for image manipulation. {Co-current DisDiff~\cite{yang2023disdiff} proposes a vector-wise method to express more information compared with other scalar-based methods~\cite{burgess2018understanding,lin2019infogancr,kwon2023diffusion}. However, they}
uses gradient fields as an inductive bias at all time steps to achieve DRL, which relies on an external heuristic hand-craft disentangling loss, lacking a self-driven learning ability and is hard to optimize. 
In contrast, our CL-Dis builds a closed-loop DRL framework with a mutual-promoting mechanism driven by a two-branch complementary diffusion and $\beta$-VAE architecture, encouraging stronger disentangled feature auto-learning.

\subsection{Knowledge Distillation and Feature Feedback}



Our closed-loop CL-Dis is bridged by these two techniques. Knowledge distillation (KD) is originally developed for model compression and acceleration~\cite{cheng2017survey,gou2021knowledge,wang2018progressive}, or used in transfer learning~\cite{ji2021show,ahn2019variational}. Feature feedback~\cite{poulis2017learning,dasgupta2018learning,katakkar2021practical,dasgupta2020robust,luo2023hierarchical} enhances the robustness of learning systems by using explainable information, which is gaining more and more attention in AI alignment~\cite{ji2023ai}, RLHF~\cite{knox2011augmenting,dai2023safe}, and embodied AI~\cite{savva2019habitat,driess2023palm}. For instance, ControlVAE~\cite{shao2020controlvae} achieves much better reconstruction quality with a variant of the proportional-integral-derivative (PID) control, using the output KL-divergence as feedback. Following this idea, we explore taking the strong generation capability of diffusion models as information-increasing feedback to automatically and adaptively enhance disentangled feature learning.

\subsection{Latent Semantics Discovery (LSD)}

This task emerges to move around the latent code (of GAN~\cite{voynov2020unsupervised,cherepkov2021navigating,song2023householder}, Diffusion~\cite{kwon2023diffusion}, etc), such that only one factor varies during the traversal generation. They mostly rely on human annotations (i.e., segmentation masks, attribute categories, 3D priors, and text descriptions) to define the semantic labels~\cite{goetschalckx2019ganalyze,shen2020interpreting,plumerault2020controlling,li2021surrogate,chen2022exploring,shi2022semanticstylegan}. Unsupervised LSD~\cite{voynov2020unsupervised,shen2021closed,song2023latent} often learn a set of directions or a classifier/matrix (e.g., Hessian, Jacobian regularization, etc.) to identify latent semantics. Wu et al.~\cite{wu2023uncovering} find semantics in
the stable diffusion model by partially changing the text embeddings. However, these methods are not strictly equivalent to DRL, most of them just rely on already disentangled (text) features to get coherent traversal generation results.




\section{Background and Motivation}

\subsection{Understanding disentangling in $\beta$-VAE}

Due to the poor disentanglement of vanilla VAE on complex data, $\beta$-VAE~\cite{higgins2016beta} adds explicit inductive bias to strengthen the independence constraint of the variational posterior distribution $q_{\phi}(\mathbf{z}|\mathbf{x})$ with a ${\beta}$ penalty coefficient in ELBO:

{\footnotesize
\begin{align}
\mathcal{L}(\theta, \phi)=
&\mathbb{E}_{q_{\phi}(\mathbf{z} | \mathbf{x})}\big[\log p_{\theta}(\mathbf{x} | \mathbf{z})\big]- \beta D_{K L}\big(q_{\phi}(\mathbf{z} | \mathbf{x}) \| p_{\theta}(\mathbf{z})\big),
\label{eq:beta-vae}
\end{align}
}\normalsize

\noindent when ${\beta}$=1, ${\beta}$-VAE degenerates to the original VAE, and larger ${\beta}$ encourages more disentangled representations but harms the performance of reconstruction. Unfortunately, it is practically intractable to obtain the optimal $\beta$ to balance the trade-off between reconstruction quality and the disentangling capability~\cite{kumar2017variational,chen2018isolating}. Thus, Burgess et al.~\cite{burgess2018understanding} understand ${\beta}$-VAE from the perspective of information bottleneck theory, and propose improving disentangling in ${\beta}$-VAE with controlled information capacity increase:

{\footnotesize
\begin{align}
\mathcal{L}(\theta, \phi)=&\mathbb{E}_{q_{\phi}(\mathbf{z} \mid \mathbf{x})} \log p_{\theta}(\mathbf{x} | \mathbf{z})-\beta\big|D_{K L}\big(q_{\phi}(\mathbf{z} | \mathbf{x}) \| p_{\theta}(\mathbf{z})\big)-C\big|,
\label{eq:understanding_bae}
\end{align}
}\normalsize

\noindent where $C$ gradually increases from $0$ to a value large enough during the training to guarantee the expressiveness of latent representations, thus improving disentangling. However, such a hand-craft increasing strategy is hard to control and might hurt disentangling performance, especially for various scenarios {on different datasets}. \textit{Thus, in our work, we learn from this idea but upgrade it by exploiting the essence of the Diffusion Probabilistic Model (DPM) in automatically increasing information capacity~\cite{pierobon2012capacity,lee2023revised} for disentangling enhancement.}

\subsection{DPM and Diffusion Autoencoder (Diff-AE)}



Here, we first take DDPM~\cite{ho2020denoising} as a vanilla DPM example for illustration, which adopts a sequence of fixed variance distributions $q(x_t|x_{t-1})$ as the noise-adding forward process to collapse the image distribution $p(x_0)$ to $\mathcal{N}(0,I)$:
\begin{equation}
     q(x_t|x_{t-1}) = \mathcal{N}(x_t;\sqrt{1-\beta_t}x_{t-1}, \beta_tI).
\end{equation}
The reverse process is fitting by other distributions of $\theta$, and it is a progressively increasing process of information capacity (i.e., entropy decreasing) as we mentioned before:
\begin{equation}
     p_\theta(x_{t-1}|x_t) = \mathcal{N}(x_t;\mu_\theta(x_t,t), \sigma_tI), 
     \label{eq:diff_reverse}
\end{equation}
where $\mu_\theta(x_t,t)$ is parameterize by a Unet $\epsilon_\theta(x_t,t)$. Its optimization is to minimize the variational upper bound of negative log-likelihood $\mathcal{L}_\theta = \mathop{\mathbb{E}}_{x_0,t,\epsilon} \|\epsilon - \epsilon_\theta(x_t,t)\|$.


Furthermore, Diff-AE~\cite{preechakul2022diffusion} is proposed for representation learning based on DPM and Autoencoder. It uses a learnable \textbf{semantic encoder} for discovering the high-level meaningful semantics $\mathbf{z_{sem}} = E_{sem}(x_0)$, and a DPM-based \textbf{decoder} $p(x_{t-1}|x_t, \mathbf{z_{sem}})$ that is conditioned on $\mathbf{z_{sem}}$ for reconstruction. However, the latent representation $\mathbf{z_{sem}}$ learned by Diff-AE does not explicitly respond to the underlying factors of the data, or say, its disentangling potential has not been fully unleashed.

\begin{figure*}
    \centering
    \vspace{-5mm}
    \includegraphics[width=0.92\linewidth]{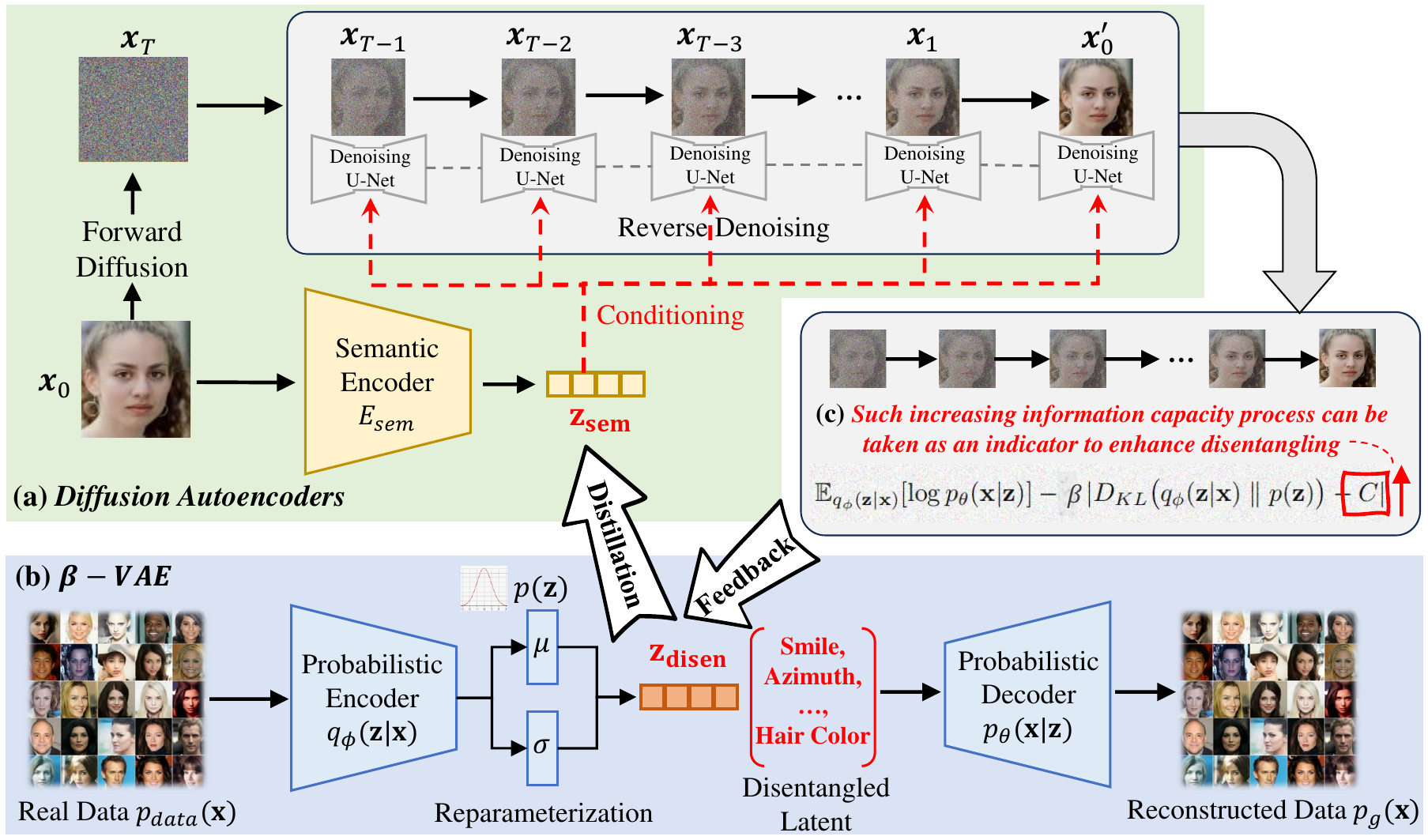}
    \vspace{-2mm}
    \caption{Pipeline of the proposed CL-Dis. We explore using \textbf{(a) diffusion-based autoencoder (Diff-AE)} as a backbone while resorting to \textbf{(b) $\beta$-VAE} as a co-pilot via knowledge distillation to help $E_{sem}$ extract a semantic disentangled representation $\mathbf{z_{sem}}$ of an input image $x_0$ (which is taken as condition for diffusion-based generation). To further facilitate feature disentangling, we further take \textbf{(c) the reverse diffusion process as an increasing information capacity indicator feedback} to, in turn, enhance the disentanglement of feature $\mathbf{z_{disen}}$ in $\beta$-VAE. As a result, Diff-AE and $\beta$-VAE are interconnected as a closed-loop cycle system with VAE-latent distillation and diffusion-wise feedback controlling, achieving a win-win effect and well-disentangled fine-controllable representation learning.}
    \vspace{-6mm}
    \label{fig:pipeline}
\end{figure*}

\section{Methodology: Closed-Loop DRL (CL-Dis)}

Based on the analysis above, we propose to leverage VAE and diffusion to unleash their strengths for a mutually promoting DRL. As shown in Figure~\ref{fig:pipeline}, our CL-Dis is a dual-branch cycle framework, involving multiple iterative phases with (a) diffusion autoencoder (Diff-AE), (b) $\beta$-VAE, and the core (c) knowledge distillation \& feature feedback process. In Section~\ref{4.1}, we first present the formulation of closed-loop DRL and the overview
of CL-Dis. After that, we introduce different phases of CL-Dis, including Diff-AE pre-training, $\beta$-VAE pre-training, and knowledge distillation \& feature feedback in Section~\ref{4.2}. Then, we propose a self-supervised Navigation strategy to identify disentangled semantic meanings and a new well-designed DRL metric (Section~\ref{4.3}) for the final disentanglement measurement.

\subsection{Problem Formulation and Overview of CL-Dis}
\label{4.1}






Given a practical dataset $\mathcal{D}$ derived by $N$ underlying factors $\mathcal{C} =\{1,2,\dots,N\}$, the target of disentanglement for our CL-Dis is to learn a semantic encoder $E_{sem}$, for each factor $c \in \mathcal{C}$, when varying its latent value, its reconstructed generation results will have the corresponding changes. 

In Figure~\ref{fig:pipeline}, our CL-Dis takes Diff-AE as the backbone, which has exactly a semantic encoder $E_{sem}$ to encode the raw data $x_0$ into the semantic representations $\mathbf{z_{sem}}$ for conditioning the diffusion reverse distributions $p(x_{t-1}|x_t, \mathbf{z_{sem}})$. To empower $E_{sem}$ disentangling capability, we additionally construct a $\beta$-VAE as co-pilot to help $\mathbf{z_{sem}}$ become a semantic \emph{disentangled} representation. Given a dataset $\mathcal{X}$ from a distribution $p_{data}(\mathbf{x})$, a standard $\beta$-VAE framework~\cite{higgins2016beta} is built and $\phi$, $\theta$ parametrize the VAE encoder $q_{\phi}(\mathbf{z}|\mathbf{x})$ and the decoder $p_{\theta}(\mathbf{x}|\mathbf{z})$, respectively. The prior $p(\mathbf{z})$ is typically set to the isotropic unit Gaussian $\mathcal{N}(0, 1)$, using a differentiable ``reparametrization trick ($z = \mu + \sigma$)''~\cite{kingma2013auto}. With the objective of Eq.~\ref{eq:beta-vae}, we get a preliminary disentangled representations $\mathbf{z_{disen}}$. Then, we use a distillation loss $\mathcal{L}_{dt}$ to transfer such disentangling capability to the semantic latent $\mathbf{z_{sem}}$ of Diff-AE,
{\small
\begin{equation}
  \mathcal{L}_{dt}=D_{KL}\big(\hspace{1mm}\mathbf{z_{sem}} \hspace{1mm} \| \hspace{1mm} \mathbf{z_{disen}} \hspace{1mm} \big) = \sum \mathbf{z_{sem}} \cdot \log (\frac{\mathbf{z_{sem}}}{\mathbf{z_{disen}}} ),
\label{eq:distill}
\end{equation}}
\noindent As discussed before, gradually increasing the information capacity of the VAE latent with a controlled way could improve disentangling~\cite{burgess2018understanding}. The reverse diffusion process of Diff-AE (Eq.~\ref{eq:diff_reverse}) is exactly a procedure of information capability increasing. Thus, we take it as an indicator (see Figure~\ref{fig:pipeline}(c)) in turn to enhance $\beta$-VAE disentangling by adaptively adjusting the objective of Eq.~\ref{eq:understanding_bae}, where we change the hand-craft $C$ to a variable that can automatically change in the optimization (Eq.~\ref{eq:fd_loss}). Such a mutually promoting mechanism forms a closed-loop optimization flow.



\subsection{Training of the Whole CL-Dis Framework}
\label{4.2}

\subsubsection{Phase-1: Diff-AE and $\beta$-VAE Pre-training}
\label{phase-1}

 
Following~\cite{preechakul2022diffusion}, the entire Diff-AE model is pre-trained on the various datasets~\cite{karras2019style,karras2017progressive,yu2015lsun} to cover more semantic factors in the latent space of $\mathbf{z_{sem}}$. But at this phase, $\mathbf{z_{sem}}$ is still far from well-disentangled, which is just taken as a kind of coarse condition for guiding diffusion generation~\cite{yang2023disdiff}.

Moreover, the co-pilot $\beta$-VAE is also pre-trained on the corresponding datasets (i.e., if Diff-AE was trained on the facial data, $\beta$-VAE did the same), following the objective of Eq.~\ref{eq:beta-vae}, its latent $\mathbf{z_{disen}}$ has already enabled a preliminary disentangling capability, i.e., can distinguish facial features of \textit{smile, azimuth, hair color, etc} (Figure~\ref{fig:pipeline}) to some degree. 

\vspace{-2mm}
\subsubsection{{\small Phase-2: Knowledge Distillation {\footnotesize \&} Diffusion Feedback}}

To transfer the preliminary disentangling capability of $\mathbf{z_{disen}}$ into $\mathbf{z_{sem}}$, we use a distillation loss shown in Eq.~\ref{eq:distill} to smoothly bridge Diff-AE and $\beta$-VAE. But this interaction is currently one-way, which is still limited by the intrinsic shortcomings of $\beta$-VAE discussed after Eq.~\ref{eq:beta-vae} that it is hard to balance a trade-off and thus hurt the final performance.


Therefore, we propose a novel diffusion feedback-driven loss $\mathcal{L}_{fd}$ to further facilitate the disentanglement of the representations ($\mathbf{z_{disen}}$, $\mathbf{z_{sem}}$) and break the optimization bottleneck of the entire CL-Dis framework. Basically, inspired by the work of understanding $\beta$-VAE~\cite{burgess2018understanding} that modified the vanilla objective by adding a hand-craft increasing controller $C$ (as shown in Eq.~\ref{eq:understanding_bae}), we propose to leverage the reverse diffusion of Diff-AE that adaptively increases information
capacity to build constraints to replace the original format. Formally, we make the original $C$ auto-changeable, and control its value dynamically according to the information capacity of the reverse diffusion process as follows:

{{\footnotesize
\begin{equation}
    C_{dyn} = f({E_{x_t}}) = \begin{cases}
  C_{base} \cdot (\frac{E_{x_0}}{E_{x_t}}),& \text{ if } 0<f({E_{x_t}})<C_{max} \\
  C_{max}, &\text{ if } f({E_{x_t}}) \geq C_{max}
\end{cases}
\end{equation}}\normalsize
{\footnotesize
\begin{equation}
E_{{x_0}} = -\sum_{ p_{{x_0}} \in \mathbb{R}} p_{{x_0}}logp_{{x_0}}, 
\hspace{2mm}
E_{{x_t}} = -\sum_{p_{{x_t}} \in \mathbb{R}} p_{{x_t}}logp_{{x_t}}, 
\end{equation}}\normalsize}

\noindent where $E_{x_0}$ and $E_{x_t}$ denote the information entropy of the static target image (i.e., Ground Truth) $x_0$ and the intermediate predicted diffusion denoising image result $x_t$, respectively. 
The probability values of $p_{x_0}, p_{x_t}$ are calculated by flattening and probabilizing the values of the image matrices across the dimension of $\mathbb{R}^{C\times H\times W}$. Besides, $C_{base}, C_{max}$ are the original empirical start-point and the maximum upper-bound according to~\cite{burgess2018understanding}, and their influences are studied in supplementary. We can intuitively see that with the denoising diffusion processing, the entropy $E_{x_t}$ of the predicted image result $x_t$ will gradually drop (i.e., information capacity increasing), and thus the ratio of $\frac{E_{x_0}}{E_{x_t}}$ will increase to raise the influence of $C_{dyn}$. So, the objective of $\beta$-VAE changes from Eq.~\ref{eq:understanding_bae} to the following $\mathcal{L}_{fd}$,

{\footnotesize
\begin{align}
\mathcal{L}_{fd}(\theta, \phi, C_{dyn} ; \mathbf{x}, \mathbf{z})=&\mathbb{E}_{q_{\phi}(\mathbf{z} \mid \mathbf{x})} \log p_{\theta}(\mathbf{x} | \mathbf{z})-  \nonumber \\
&\beta\big|D_{K L}\big(q_{\phi}(\mathbf{z} | \mathbf{x}) \| p_{\theta}(\mathbf{z})\big)-C_{dyn}\big|,
\label{eq:fd_loss}
\end{align}
}\normalsize

\noindent the above physical meaning is that: when KL-divergence drops to a certain extent (i.e., disentangling ability encounters bottleneck in a naive VAE latent), the dynamic controller $C_{dyn}$ will counteract this change by increasing the value of the second item of Eq.~\ref{eq:fd_loss}. Intuitively, $C_{dyn}$ that is controlled by diffusion feedback gradually highlights the penalty for KL-divergence, thus improving the latent disentangling. Under this closed-loop optimization flow (see Figure~\ref{fig:pipeline}(c)), the gradually increasing information capacity of diffusion adaptively enhances the disentanglement of latent $\mathbf{z_{disen}}$, so as to further benefit the semantic code of $\mathbf{z_{sem}}$. Please see supplementary for more theoretical analysis.


\vspace{-3mm}
\subsubsection{Phase-3: Semantics Directions Navigation}

\begin{figure}
    \centering
    \includegraphics[width=\linewidth]{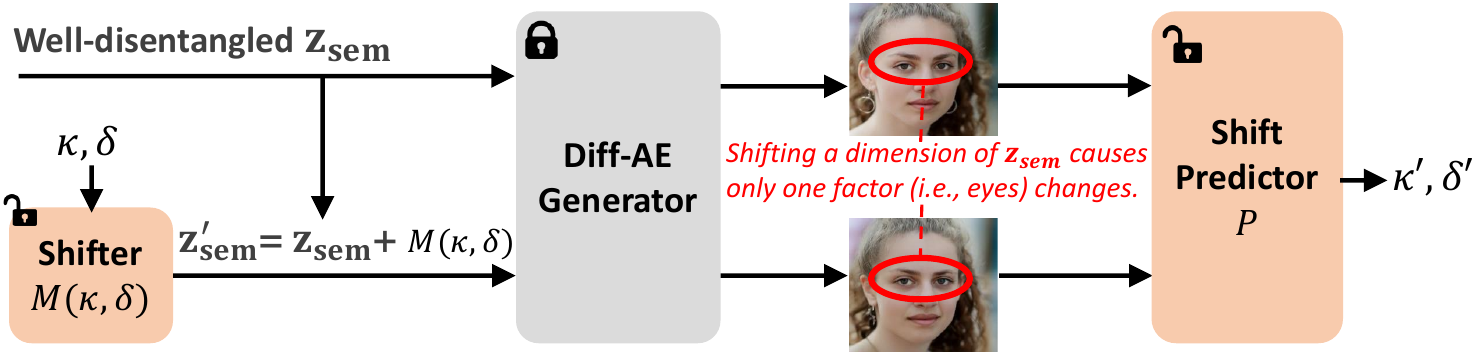}
    \vspace{-6mm}
    \caption{A self-supervised Navigation strategy is to discover the interpretable shifts in the well-disentangled space of $\mathbf{z_{sem}}$. A training sample consists of two images, produced by the Diff-AE with original and shifted inputs. The images are given to a predictor $P$ that predicts the direction index $k$ and shift magnitude $\delta$.}
    \vspace{-6mm}
    \label{fig:navigation}
\end{figure}

After getting the well-disentangled representations of $\mathbf{z_{sem}}$, the next step is to identify its semantic meanings. Shifting a certain dimension of the well-disentangled condition $\mathbf{z_{sem}}$ will result in a generated result that only the corresponding property/factor changes. Thus, we adopt an outer navigation branch to identify the traversal directions in a global view. Specifically, we employ a learnable matrix $M$ (see Figure~\ref{fig:navigation}) to manually append a shift on $\mathbf{z_{sem}}$. The shifted code $\mathbf{z_{sem}'}$, together with the original $\mathbf{z_{sem}}$, is fed into our Diff-AE decoder to generate a pair of images, following the reverse conditional diffusion process of $p(x_{t-1}|x_t, \mathbf{z_{sem}})$. A learnable predictor $P$ is then devised to predict the shift (i.e., semantic direction) according to such generated paired images, with a reconstruction loss. In this way, we clearly identify the inner disentangled semantics directions in our CL-Dis for further fine-controllable generations.




\subsection{New Disentanglement Metric}
\label{4.3}

The conditional diffusion-based generation in our CL-Dis defines a clear mapping between the disentangled representations and the generated image results, so we can quantitatively measure the performance of disentanglement with the generation difference. Following this idea,  we quantify this difference by using an optical flow tracking technique~\cite{teed2020raft}: Given a well-disentangled representation $\mathbf{z_{sem}^a}$ and its shift variant $\mathbf{z_{sem}^{a'}}$, we calculate the optical flow map $F$ between their corresponding generated images. Then, we use a continuously changed hyper-parameter as a dynamic threshold to distinguish these ``\emph{changed characteristics}'' caused by shifted factors from the rest ``\emph{frozen ones}''. The average ratio of these two parts actually reflects the disentanglement performance, the smaller the better. Due to the space limitations, more operation details are in supplementary.

\section{Experiment}




\subsection{Implementation Detail and Experiment Setting}

For the backbone of Diff-AE, we take pre-trained weights in~\cite{preechakul2022diffusion} as the initial parameters with Adam optimizer and the learning rate of 1e-4. For the co-pilot $\beta$-VAE, we adopt the same Adam optimizer and learning rate. For the navigation branch, we refer to the GAN-based editing methods~\cite{voynov2020unsupervised,cherepkov2021navigating} for a fine-controllable generation. Differently, our CL-Dis has already in advance got the well-disentangled latent $\mathbf{z_{sem}}$. The whole CL-Dis framework is trained on 2 NVIDIA V100 GPUs with a batch size of 32. In inference, the co-pilot $\beta$-VAE is discarded for efficiency. The dimension of $\mathbf{z_{sem}}$ is set to 512 in our framework. To evaluate disentanglement, we conduct experiments on both generation \& perception tasks, on synthetic \& real data.

\noindent\textbf{Datasets \& Baselines.} For the image generation task, we conduct experiments on FFHQ~\cite{karras2019style}, CelebA~\cite{karras2017progressive}, LSUN Horse~\cite{yu2015lsun}, LSUN Cars~\cite{yu2015lsun}, Shape3D~\cite{kim2018disentangling}. 
We compare our proposed CL-Dis with VAE-based, GAN-based, and Diffusion-based baselines, including  
$\beta$-VAE~\cite{higgins2016beta},
FactorVAE~\cite{kim2018disentangling},
Beta-TCVAE~\cite{chen2018isolating},
GAN Edit v1~\cite{voynov2020unsupervised}, GAN Edit v2~\cite{cherepkov2021navigating}, InfoGAN-CR~\cite{lin2019infogancr}, Asyrp~\cite{kwon2023diffusion}, GANspace (GS) ~\cite{harkonen2020ganspace}, NaviNeRF~\cite{xie2023navinerf}, DisCo~\cite{ren2021learning}, Diff-AE~\cite{preechakul2022diffusion} and vector-based DisDiff~\cite{yang2023disdiff}.

For the perception task, we just choose the classic face recognition for instantiation due to limited space and conduct experiments on CASIA-Webface \cite{CASIA-Webface} and VGGFace2 \cite{vggface2} in comparison with the baseline of LFW~\cite{lfw}.

\noindent\textbf{Evaluation Metrics.} To alleviate errors, we have multiple runs for each method. We use 4 representative metrics of Factor-VAE score~\cite{kim2018disentangling}, DCI~\cite{eastwood2018framework}, FID~\cite{heusel2017gans} and classification accuracy in face recognition, and our additionally proposed new metric for validation. Note that our CL-Dis belongs to dimension-/scalar-based DRL methods, so we just discuss but do not directly compare with many vector-based ones~\cite{denton2017unsupervised,tran2017disentangled,liu2021activity,yang2023vector,chen2023disenbooth} in the next experiment sections. 

\subsection{Comparisons with SOTA DRL Methods}


\begin{table}[!ht] 
\begin{minipage}[b]{0.60\linewidth}
\begin{center}
\renewcommand\tabcolsep{15.0pt}
\caption{Quantitative comparison results on CelebA dataset.}
\scriptsize

\begin{tabular}{cccccc}
\toprule
Methods & FID \\
\midrule
VAE~\cite{kingma2013auto}             &  132.80 \\
$\beta$-VAE~\cite{higgins2016beta}    &  136.23 \\
$\beta$-TCVAE~\cite{chen2018isolating}   &  139.07 \\
FactorVAE~\cite{kim2018disentangling}    &  134.52 \\
\midrule

CL-Dis (Ours)   & \textbf{6.54} \\
\bottomrule
\end{tabular}

\vspace{0pt}
\label{FID}
\end{center}
\end{minipage}
\hfill
\begin{minipage}[b]{0.3\linewidth}
    \begin{center}
    \renewcommand\tabcolsep{3.0pt}
\caption{Quantitative comparison results on FFHQ dataset.}
\scriptsize

\begin{tabular}{cccccc}
\toprule
Methods & FID  \\
\midrule
NaviNeRF~\cite{xie2023navinerf} & 13.00  \\  
CL-Dis (Ours) & \textbf{12.15}  \\
\bottomrule
\end{tabular}

\vspace{0pt}
\label{FID2}
\end{center}
\end{minipage}
\vspace{-10pt}
\end{table}

\noindent\textbf{Quantitative Results.} 
Table~\ref{FID} and Table~\ref{FID2} report the results of Frechet Inception Distance (FID)~\cite{heusel2017gans} to measure the quality of the generated images. We see that our CL-Dis
outperforms other typical disentangled methods in generation quality, especially on the real-world face images of CelebA. 

Besides, the quantitative comparison results of disentanglement under different metrics are shown in Table~\ref{disqnti}. As shown in the table, {CL-Dis outperforms other competitors,} demonstrating the
method’s superior disentanglement ability. Compared with
the VAE-based methods, these methods suffer from the
the trade-off between generation and disentanglement~\cite{higgins2016beta,burgess2018understanding}, but CL-Dis does not have this problem due to its automatic optimization mechanism. As for the GAN-based methods are typically limited by the latent space of GAN~\cite{yang2023disdiff}. The diffusion-based DisDiff is a vector-wise method, in which representations $\{z^c\}$ could express more information than our scalar-based ones $\mathbf{z_{sem}}$, but our CL-Dis still outperforms DisDiff on Shape3D. And, our scalar-based disentangled representations are more light, generic, robust, and easy to apply to different scenarios.

\begin{table}[t]
\vspace{-0pt}
\small{
\caption{{Comparisons of disentanglement on the FactorVAE score and DCI disentanglement metrics (mean $\pm$ std, higher is better). We can see that our scalar-based CL-Dis even achieves comparable results with the co-current factor-based vector-wise DisDiff~\cite{yang2023disdiff}}.}
\label{disqnti}
\vspace{-10pt}
\begin{center}
\setlength{\tabcolsep}{0.3mm}{
\begin{tabular}{ccccc}
\toprule
\multicolumn{2}{c}{\multirow{2}{*}{\textbf{Method}}} &
\multicolumn{2}{c}{Shapes3D} &  \\
\cmidrule(lr){3-4}
&   & FactorVAE score & DCI \\
\midrule

\multirow{2}*{\textbf{VAE-based:}} & FactorVAE~\cite{kim2018disentangling}  & 0.840 ± 0.066 & 0.611 ± 0.082 \\

& $\beta$-TCVAE~\cite{chen2018isolating} & 0.873 ± 0.074 & 0.613 ± 0.114   \\

\midrule

{\textbf{GAN-based:}}
& InfoGAN-CR~\cite{lin2019infogancr}  &0.587 ± 0.058 &0.478 ± 0.055  \\

\midrule

\multirow{3}{*}{\begin{tabular}[c]{@{}l@{}}\textbf{Pre-trained}\\ \textbf{GAN-based:}\end{tabular}}
& GAN Edit v1~\cite{voynov2020unsupervised}  & 0.805 ± 0.064 & 0.380 ± 0.062  \\

& GANspace~\cite{harkonen2020ganspace}  & 0.788 ± 0.091 & 0.284 ± 0.034  \\

& DisCo~\cite{ren2021learning}  & 0.877 ± 0.031 & 0.708 ± 0.048   \\

\midrule

\multirow{2}*{\textbf{Diffusion-based:}}
& DisDiff~\cite{yang2023disdiff}  & 0.902 ± 0.043 & 0.723 ± 0.013 \\

& CL-Dis (Ours)  & \textbf{0.952 ± 0.017}  & \textbf{0.731 ± 0.045}  \\

\bottomrule
\end{tabular}}
\vspace{-10pt}
\end{center}}
\end{table}









\begin{figure}
    \centering
\includegraphics[width=\linewidth]{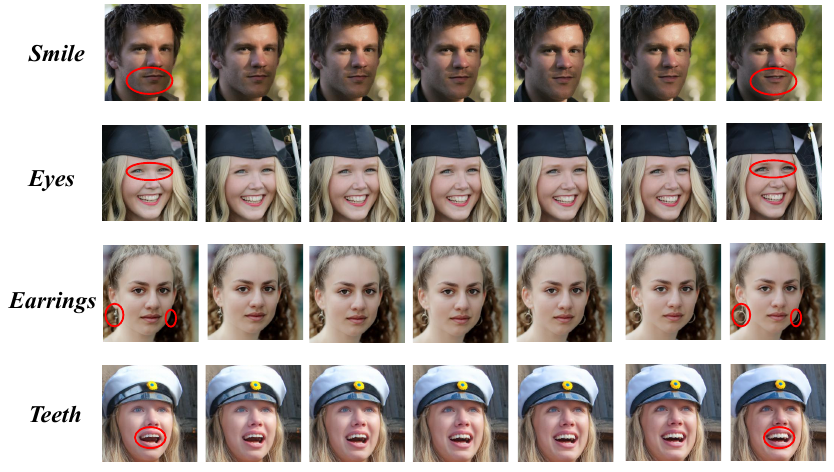}
    \vspace{-18pt}
    \caption{Fine-grained Disentanglement Results. The qualitative results present the results of attribute manipulation, which demonstrate the semantic manipulation on the FFHQ dataset. To clearly show the isolated and ``disentangled'' effects, we use red circles to mark the corresponding changed regions.}
    \label{FineDis}
    \vspace{-15pt}
\end{figure}

\begin{figure}
    \centering
    \includegraphics[width=\linewidth]{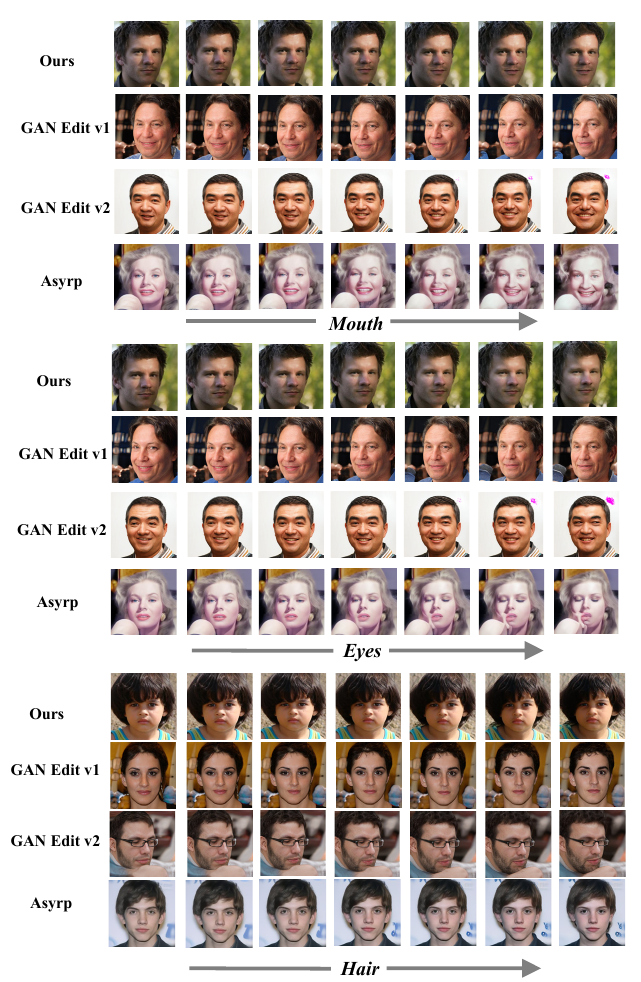}
    \vspace{-18pt}
    \caption{Comparison of the ``mouth'', ``eyes'' and ``hair" attributes with different methods on the FFHQ dataset. {Since different methods pre-train on different datasets, we thus compare the disentanglement with different images according to the previous methods ~\cite{voynov2020unsupervised,cherepkov2021navigating,kwon2023diffusion}}.
    }
    \label{compare}
    \vspace{-20pt}
\end{figure}



\noindent\textbf{Qualitative Results.} 
{Unlike most previous papers that only test on the toy synthetic datasets~\cite{ higgins2016beta,kim2018disentangling,chen2018isolating}, our CL-Dis can be naturally applied to real images.} Figure~\ref{FineDis} shows the effectiveness of our method on various datasets. To clearly
show the isolated and ``disentangled'' effects, we use red circles to
mark the corresponding changed regions. We can observe that CL-Dis can disentangle and change the \emph{isolated} attributes continuously in an unsupervised manner, such as \emph{serious expression $\rightarrow$ smile, closing eyes $\rightarrow$ opening eyes, w/o earrings $\rightarrow$ w/ earrings, etc.} 

Moreover, to validate the superiority of CL-Dis in the disentanglement capability, Figure~\ref{compare} {also provides results for changing human facial attributes to different characteristics along certain directions like ``Mouth'', ``Eyes'' and ``hair"}, in comparison with three baselines of \textit{GAN Edit v1}~\cite{voynov2020unsupervised}, \textit{GAN Edit v2}~\cite{cherepkov2021navigating}, and \textit{Asyrp}~\cite{kwon2023diffusion}. We can see that other methods like Asyrp are prone to generate artifacts (bottom row of Figure~\ref{compare}), and their editing results are not well-disentangled. For example, the whole face content has changed for the schemes of \textit{GAN Edit v1}~\cite{voynov2020unsupervised} and \textit{GAN Edit v2}~\cite{cherepkov2021navigating}). In contrast, our method consistently keeps the identity while just changing the disentangled properties, achieving an ``isolated and disentangled” effect.

\begin{figure}
    \centering
    \includegraphics[width=\linewidth]{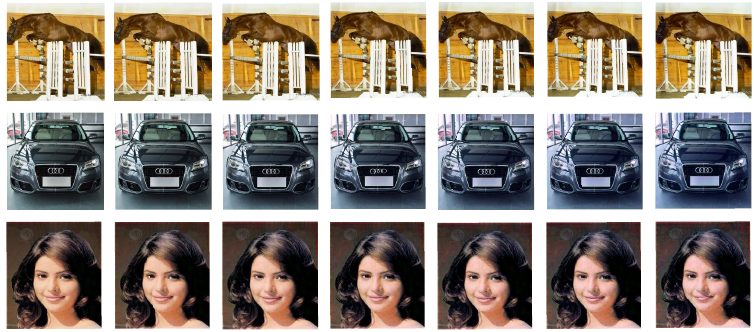}
    \vspace{-18pt}
    \caption{
    Generalization evaluation on extensive datasets. We visualize the results of shifting the latent $\mathbf{z}_{sem}$ with our proposed method along the common ``Color'' direction on LSUN Horse (upper row), LSUN Cars (middle row), and CelebA (lower row) datasets, respectively.}
    \label{generalization}
    \vspace{-18pt}
\end{figure}

To demonstrate the generalization ability of our CL-Dis, we also use different datasets and different backbones (supplementary) for validation. As shown in Figure~\ref{generalization}, even for cars, horses, and buildings, CL-Dis still obtains a group of fine-grained disentanglement results, where only the ``color'' attribute changes across the traversal direction, proving the disentanglement generalization capability.

\begin{figure*}
    \centering
    \vspace{-20pt}
    \includegraphics[width=0.9\linewidth]{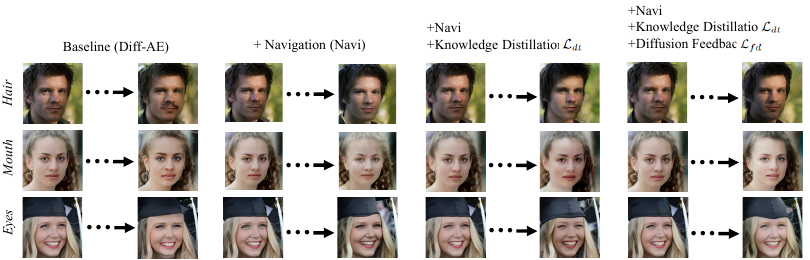}
    \vspace{-10pt}
    \caption{Ablation study on the FFHQ dataset. We evaluate the effectiveness of different components in our method by shifting different attributes, including ``Hair'', ``Mouth'' and ``Eyes''. We can see that as more elements are added, both the disentanglement performance and the generation quality improve: the first column exists artifacts; the second and third columns change other contents except shift attributes.}
    \label{ablation}
    \vspace{-15pt}
\end{figure*}

\subsection{Ablation Study and More Analysis}

\noindent\textbf{Effectiveness of Different Elements.} To analyze the effectiveness of the proposed parts in our closed-loop disentanglement framework, we perform an ablation to study the influence of Semantics Navigation, Knowledge Distillation $\mathcal{L}_{dt}$, and Diffusion Feedback $\mathcal{L}_{fd}$. We take FFHQ as the dataset to conduct these ablation studies.

As shown in Figure~\ref{ablation}, due to the baseline of Diff-AE itself not having a disentanglement ability, when shifting its latent we get generation results with background artifacts. If we directly apply the semantics navigation~\cite{voynov2020unsupervised} to the baseline latent, the generation quality has been improved, but the disentanglement is still not guaranteed. For example, when
shifting along the semantic direction of ``Hair'' (The 1$^{st}$ row of Figure~\ref{ablation}), global features such as face
shape, skin texture, and hairstyle are variously entangled, resulting in a younger appearance for the man. The last two schemes of adding knowledge distillation and diffusion feedback have achieved better performance in disentanglement, particularly for the final closed-loop scheme. We can see that the final scheme together with $\mathcal{L}_{dt}$ \& $\mathcal{L}_{fd}$ achieves a better disentanglement effect on the semantic direction of ``Hair'', compared to that only with $\mathcal{L}_{dt}$ which still changes face characteristics in the traversal generations.










\noindent\textbf{Analysis of Auto-Driven Feedback Mechanism.} As introduced, the core contribution of this work lies in an automatically disentangled learning mechanism, achieved by a dynamic $C_{dyn}$ whose value depends on the increasing information capacity in the reverse diffusion process. To prove that, we visualize the changing curves of $C_{dyn}$ during the CL-Dis training. As illustrated in Figure~\ref{curve}(a), the value of $C_{dyn}$ gradually increases with the diffusion feedback, so as to highlight the penalty for KL-divergence item in the co-pilot $\beta$-VAE optimization and thus improve disentanglement performance of the entire framework.

\subsection{New Metric $\&$ Discrimination Task Validation}

\begin{figure}
    \centering
    \includegraphics[width=1.0\linewidth]{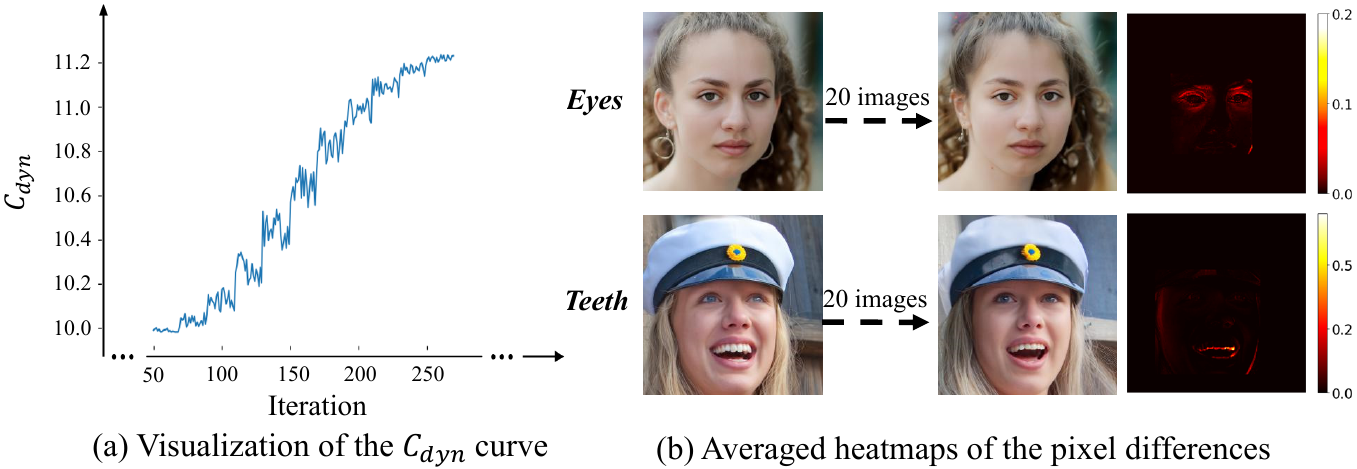}
    \vspace{-15pt}
    \caption{(a) Visualization of the curve of changeable $C_{dyn}$ at different iteration steps. $C_{dyn}$ gradually increases with the diffusion feedback to improve the latent disentangling. (b) Averaged heatmaps of the pixel differences between the original and the shifted images for the ``eyes shift" (left) and the ``teeth shift" (right), revealing a good disentanglement of ours.}
    \label{curve}
    \vspace{-15pt}
\end{figure}

\begin{figure}
    \centering
    \includegraphics[width=\linewidth]{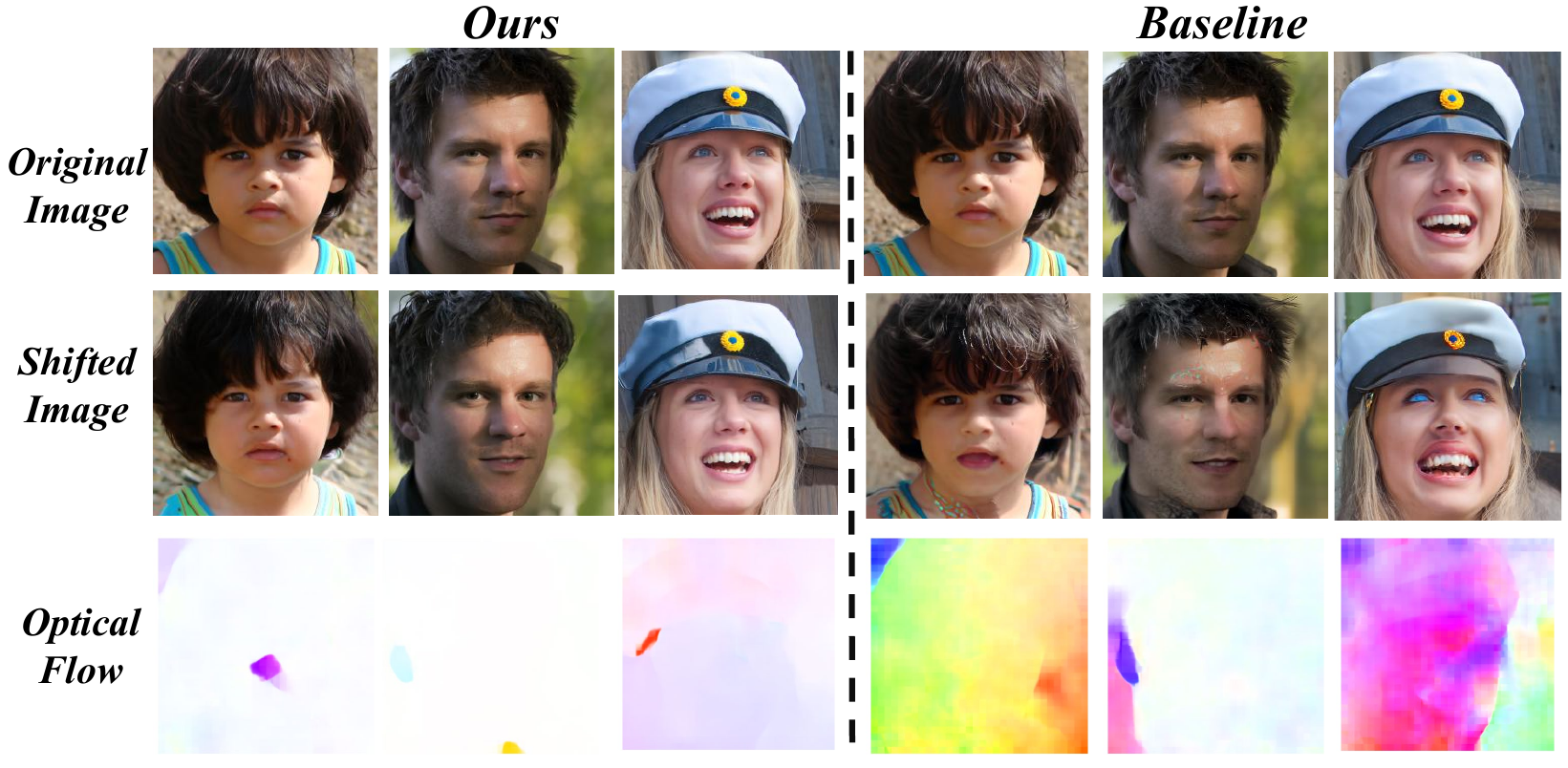}
    \vspace{-18pt}
    \caption{Evaluation on the proposed new disentanglement metric {with the ‘hair’, ‘eyes’, and ‘smile’ attributes}. From the optical flow tracking visualization, we find traversing along our disentangled representations results in an ``property isolated'' generation effect, while the baseline exhibits a bad performance with a fully changed image content.}
    \label{fig:new_metric}
    \vspace{-18pt}
\end{figure}

\noindent\textbf{New Metric Evaluation.}
{As our work belongs to the track of unsupervised representation disentanglement, not image generation or editing. The core of this task is to fully understand the latent factors of the model to enhance its fine-grained controllability.
The qualitative results shown above have illustrated that traversing along the disentangled representations results in an ``property isolated'' generation effect. To measure such effect quantitatively and show our method could progressively change the disentangled properties while keeping others frozen,}
we propose a new metric (Sec.~\ref{4.3}) through the lens of optical flow~\cite{teed2020raft}. As illustrated in Figure ~\ref{fig:new_metric}, the optical flow visualization between the original and shifted image pairs on FFHQ dataset~\cite{karras2019style}, shows that our proposed CL-Dis method demonstrates effective disentanglement, where shifting a single dimension of the latent space results in isolated property changes in the generated output. {In contrast, the baseline approach (original Diff-AE) changes the entire background content of the image.}


Furthermore, we perform normalization on the optical flow map and utilize a dynamic threshold to classify pixels as either exceeding or not exceeding it, thereby obtaining the ratio. We make statistics based on the 100 sets, and calculate the average ratio using a static threshold of 0.5, and get the average value. The baseline quantitatively achieves 0.1606 in this new metric, while our method is only 0.0177 (the lower the better), indicating that our method has better disentanglement performance that only isolated property changes when shifting across certain dimensions.

\noindent\textbf{Locality of visual effects.}
To illustrate that the navigation along the semantic directions results in an isolated and ``disentangled'' effect, for a particular direction, we compute the per-pixel differences averaged over the generation results with 20 shift magnitudes from a uniform grid in a range. Figure~\ref{curve}(b) shows the averaged heatmap for the ``Eyes'' and ``Mouth'' directions. We can see that our method achieves a good disentanglement, where shifting ``Eyes'' just affects the eyes-related image content, and shifting ``Mouth'' just affects the mouth-related image content.

\noindent\textbf{Discrimination Task Validation.}
To validate whether the good disentanglement achieved by our CL-Dis could benefit the discrimination task, we conducted face recognition experiments on two large well-known datasets, CASIA-Webface \cite{CASIA-Webface} and VGGFace2 \cite{vggface2}. We use the semantics encoder $E_{sem}$ of CL-Dis as an off-the-shelf feature extractor and introduce two additional learnable linear layers for fine-tuning. Following the same protocol used in~\cite{facenet}, we perform a fair comparison in Table~\ref{tab:perception} with mean classification accuracy as a metric. We observe that the semantic encoder of our CL-Dis performs better than the baseline of learning from scratch (LFS) on the widely adopted test benchmark LFW~\cite{lfw}, and even our $E_{sem}$ trained with only 5 epochs still outperforms the LFS model with 50 epochs, which demonstrates that the well-disentangled representations learned by CL-Dis indeed benefit discrimination tasks.


\begin{table}[h]
\centering
\footnotesize
\vspace{-2mm}
\caption{Discrimination task (face recognition) validation. Our method with good disentangled representations achieves $>$2.0\% gains even with fewer (5) training epochs compared to baseline.}
\vspace{-3mm}
\setlength{\tabcolsep}{2.8mm}{
    \begin{tabular}{lcc}
        \toprule
            Dataset & CASIA-Webface (\%) & VGGFace2 (\%)  \\
             \midrule
             Baseline (50 epochs) & 89.45+-0.015 & 81.82+-0.019 \\
             Ours (5 epochs) & 91.72+-0.011 & 87.32+-0.014  \\
            Ours (50 epochs)  & \textbf{94.00+-0.785} & \textbf{87.97+-0.012} \\
        \bottomrule
    \end{tabular}}
    \label{tab:perception}
    \vspace{-5mm}
\end{table}






\vspace{-1mm}
\section{Conclusion}
\vspace{-1mm}

In this paper, we present CL-Dis, an unsupervised closed-loop disentanglement framework that achieves a strong DRL ability without relying on heavy supervision or heuristic loss constraints. CL-Dis embeds a closed-loop interconnection, where a pre-trained $\beta$-VAE empowers a preliminary disentanglement ability to diffusion model with a distillation loss, while the increasing information capability in the diffusion is taken as feedback, in turn, to progressively improve $\beta$-VAE disentangling, forming a core mutually-promoting mechanism. Besides, we also design a self-supervised strategy to navigate semantic directions in the well-disentangled latent for a controllable generation. A new metric is further introduced to measure the disentanglement effect. Experimental results indicate that CL-Dis achieves a state-of-the-art DRL performance, and its strong disentangled ability benefits both generation and discrimination tasks, especially on real-world images.

\newpage
\appendix
\twocolumn[{
\centering
 \vspace{20pt}
\section*{\Large \centering Supplementary Material}
 \vspace{30pt}
 }]

\section{Understanding the effects of our CL-Dis}

Constrained optimization is important for enabling deep unsupervised models to learn disentangled representations of the independent data generative factors~\cite{higgins2016beta}:

(1) In the $\beta$-VAE framework, this corresponds to tuning the $\beta$ coefficient in Eq.~\ref{eq:beta-vae} as follows, 
{\small
\begin{align}
\mathcal{L}(\theta, \phi, \mathbf{x}, \mathbf{z}, \beta)=
&\mathbb{E}_{q_{\phi}(\mathbf{z} | \mathbf{x})}\big[\log p_{\theta}(\mathbf{x} | \mathbf{z})\big]- \nonumber \\
&\beta D_{K L}\big(q_{\phi}(\mathbf{z} | \mathbf{x}) \| p_{\theta}(\mathbf{z})\big),
\label{eq:beta-vae2}
\end{align}
}\normalsize

\noindent $\beta$ can be taken as \textbf{a mixing coefficient} for balancing the magnitudes
of gradients from the reconstruction and the prior-matching components of the VAE lower bound
formulation during training. When $\beta$ is too low or too high the model learns an entangled latent representation due to either too much or too little capacity in the latent bottleneck~\cite{higgins2016beta}.

(2) In the \textit{understanding $\beta$-VAE} version with a controlled capacity increase strategy~\cite{burgess2018understanding}, this constrained optimization corresponds to tuning the $C$ coefficient in Eq~\ref{eq:understanding_bae} as follows, 
{\small
\begin{align}
\mathcal{L}(\theta, \phi, C ; \mathbf{x}, \mathbf{z})=&\mathbb{E}_{q_{\phi}(\mathbf{z} \mid \mathbf{x})} \log p_{\theta}(\mathbf{x} | \mathbf{z})-  \nonumber \\
&\beta\big|D_{K L}\big(q_{\phi}(\mathbf{z} | \mathbf{x}) \| p_{\theta}(\mathbf{z})\big)-C\big|,
\label{eq:understanding_bae2}
\end{align}
}\normalsize

\noindent where the model is trained to reconstruct the images with samples from the factor distribution, but with a range of different target
encoding capacities by pressuring the KL divergence to be at a controllable value, $C$. The training
objective combined maximizing the log-likelihood and minimizing the absolute deviation from $C$ by linearly increasing it from a low value to a high value over the course of training. The intuitive picture behind $C$ is that it gradually adds more latent encoding capacity, enabling progressively more factors of variation to be represented while retaining disentangling in previously learned factors.

(3) In the proposed closed-loop unsupervised representation disentanglement framework \textit{CL-Dis}, this constrained optimization is totally automatic and dynamic-adaptive. This is achieved under the mutual driving of reverse
diffusion of Diff-AE and VAE latent evolution. We make the constant $C$ in ~\cite{burgess2018understanding} become changeable $C_{dyn}$, controlling its value dynamically according to the information capacity variation in the reverse diffusion process as follows,

{\footnotesize
\begin{align}
\mathcal{L}_{fd}(\theta, \phi, C_{dyn} ; \mathbf{x}, \mathbf{z})=& \hspace{1mm} \mathbb{E}_{q_{\phi}(\mathbf{z} \mid \mathbf{x})} \log p_{\theta}(\mathbf{x} | \mathbf{z})-  \nonumber \\
&\beta\big|D_{K L}\big(q_{\phi}(\mathbf{z} | \mathbf{x}) \| p_{\theta}(\mathbf{z})\big)-C_{dyn}\big|, \\
C_{dyn} \propto {I(x_0; x_{t})} =& \sum_{} \sum_{} p(x_{0},x_t) \cdot \log\left(\frac{p(x_{0},x_t)}{p(x_0) \cdot p(x_{t})}\right) \\
s.t. \hspace{3mm} p_{\theta'}(x_{t-1}|x_t) =& \hspace{1mm} \mathcal{N}(x_t;\mu_{\theta'}(x_t,t), \sigma_tI), 
\label{eq:fd_loss2}
\end{align}
}\normalsize

\noindent where the $C_{dyn}$ is proportional to the information capability of the reverse process in diffusion (here we use the mutual information $I(x_0; x_{t})$ to represent its value), which is parameterized by ${\theta'}$ with a U-Net architecture and is a progressively increasing process of information capacity (i.e., entropy decreasing). Continuing this intuitive picture, we can imagine that
with the capacity of the information capability gradually increasing in the reverse diffusion, the dynamic $C_{dyn}$ would continue to utilize those extra indicators for an increasingly precise encoding, where a larger improvement can be obtained by disentangling different factors of variations in the latent.

\section{Implementation Details of the New Metric}

To measure the disentanglement effect quantitatively, especially under the unsupervised setting where no factor annotations are accessible, we propose a new label-free metric with the help of optical flow~\cite{teed2020raft}. The high-level idea is that we leverage optical flow to track the obviously and continually changed content to evaluate whether or not these changes are ``property isolated''. First, we calculate the optical flow between the original image and the shifted image. Then, we normalize the values of the optical flow map. After that, we apply a threshold to the optical flow. Points in the flow exceeding this threshold are considered to have a large shift magnitude (i.e., those disentangled attribute points), while those below it are considered to have a small shift magnitude (i.e., those rest frozen attribute points). We calculate the ratio of the number of flow points exceeding this threshold to those below it and use this ratio value as our new metric score. The lower the better, and which exactly corresponds to the definition of disentangled representations where a change in a single unit of the representation corresponds to a change in a single factor of variation of the data while being invariant to others. The specific algorithm is shown in Algorithm~\ref{alg:metric} and Algorithm~\ref{alg:norm}. Empirically and experimentally, we set the threshold as 0.5 for the calculation of the proposed new metric.

\begin{algorithm}
\caption{Compute New Metric Based on Optical Flow}
\label{alg:metric}
\begin{algorithmic}[1]
\Function{ComputeNewMetric}{$image$, $shifted\_image$, $threshold$}
    \State $flow \gets \Call{OpticalFlow}{image, shifted\_image}$
    \State $normalized\_flow \gets \Call{Norm}{flow}$
    \State $large\_shift\_count\ \mathcal{L} \gets 0$
    \State $small\_shift\_count\ \mathcal{S} \gets 0$
    \ForAll{$flow\_point$ in $normalized\_flow$}
        \If{$\Call{Magnitude}{flow\_point}> threshold$}
            \State $\mathcal{L} \gets \mathcal{L} + 1$
        \Else
            \State $\mathcal{S} \gets \mathcal{S} + 1$
        \EndIf
    \EndFor
    \State $metric \hspace{1mm} score \gets\ \mathcal{L} / \mathcal{S}$
    \State \Return $metric \hspace{1mm} score$
\EndFunction
\end{algorithmic}
\end{algorithm}

\begin{algorithm}
    \caption{Optical Flow Normalization}
    \label{alg:norm}
    \begin{algorithmic}[1]
    \Function{Norm}{$flow$}
    \State $max\_norm \gets \Call{Max}{\sqrt{\sum{flow^2}}}$
    \State $normalized\_flow \gets flow / max\_norm$
    \State \Return $normalized\_flow$
    \EndFunction
    \end{algorithmic}
\end{algorithm}

\section{More Experimental Results}

\subsection{Influence of $C_{base}$ and $C_{max}$ in Our CL-Dis}

\begin{table}[t]
\vspace{0pt}
\small{
\caption{Influence study with different values of $C_{base}$ and $C_{max}$ in our CL-Dis framework on the Cars3D dataset. We evaluate the disentanglement performance of different schemes using the metrics of FactorVAE score and DCI (mean $\pm$ std, higher is better).}
\label{abl_c}
\vspace{-0pt}
\begin{center}{
\renewcommand{\arraystretch}{1.0}
\setlength{\tabcolsep}{3.4mm}{
\begin{tabular}{cccc}
\toprule
$C_{base}$ & $C_{max}$ & FactorVAE score & DCI  \\
\midrule

2 & 5 & 0.689  ± 0.038  & 0.094 ± 0.009  \\  \midrule

5 & 15 & 0.745 ± 0.042   &  0.175  ± 0.012 \\  \midrule

10 & 11 & 0.762 ± 0.068  &  0.179  ± 0.014 \\

10 & 12 &  0.781 ± 0.035  & 0.182  ± 0.008 \\

10 & 25 & \textbf{0.794 ± 0.040} & \textbf{0.186 ± 0.007} \\   \midrule

20 & 50 &  0.734 ± 0.024 & 0.163 ± 0.015  \\ 


\bottomrule
\end{tabular}}}
\end{center}}
\end{table}

\begin{figure}
    \centering
    \includegraphics[width=\linewidth]{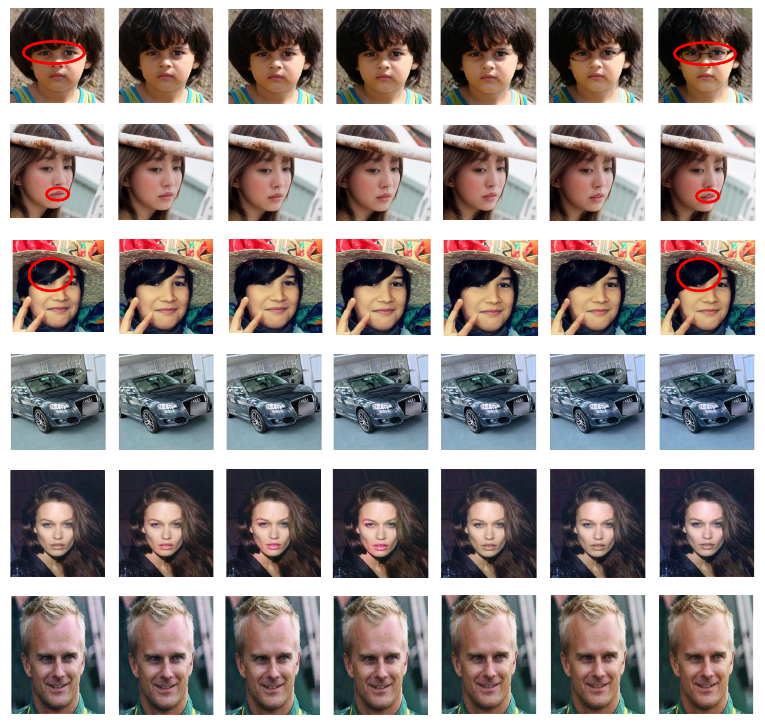}
    \vspace{0pt}
    \caption{Generalization evaluation on extensive datasets. We visualize the results of shifting the disentangled latent $\mathbf{z_{sem}}$ with our proposed method. From top to bottom, we illustrate the “property isolated” effects along \textit{glasses, mouth, hair, car color, hair color, and face illumination}, respectively.} 
    \label{supp_vis}
    \vspace{0pt}
\end{figure}

\begin{table}[t]
\vspace{-0pt}
\small{
\caption{{Comparisons of disentanglement on the FactorVAE score and DCI disentanglement metrics (mean $\pm$ std, higher is better). We can see that our scalar-based CL-Dis even achieves comparable results with the co-current factor-based vector-wise DisDiff~\cite{yang2023disdiff}}.}
\label{disqnti2}
\vspace{-10pt}
\begin{center}
\setlength{\tabcolsep}{0.3mm}{
\begin{tabular}{ccccc}
\toprule
\multicolumn{2}{c}{\multirow{2}{*}{\textbf{Method}}} &
\multicolumn{2}{c}{Shapes3D} &  \\
\cmidrule(lr){3-4}
&   & FactorVAE score & DCI \\
\midrule

\multirow{2}*{\textbf{VAE-based:}} & FactorVAE~\cite{kim2018disentangling}  & 0.840 ± 0.066 & 0.611 ± 0.082 \\

& $\beta$-TCVAE~\cite{chen2018isolating} & 0.873 ± 0.074 & 0.613 ± 0.114   \\

\midrule

{\textbf{GAN-based:}}
& InfoGAN-CR~\cite{lin2019infogancr}  &0.587 ± 0.058 &0.478 ± 0.055  \\

\midrule

\multirow{3}{*}{\begin{tabular}[c]{@{}l@{}}\textbf{Pre-trained}\\ \textbf{GAN-based:}\end{tabular}}
& GAN Edit v1~\cite{voynov2020unsupervised}  & 0.805 ± 0.064 & 0.380 ± 0.062  \\

& GANspace~\cite{harkonen2020ganspace}  & 0.788 ± 0.091 & 0.284 ± 0.034  \\

& DisCo~\cite{ren2021learning}  & 0.877 ± 0.031 & 0.708 ± 0.048   \\

\midrule

\multirow{2}*{\textbf{Diffusion-based:}}
& DisDiff~\cite{yang2023disdiff}  & 0.902 ± 0.043 & 0.723 ± 0.013 \\

& CL-Dis (Ours)  & \textbf{0.952 ± 0.017}  & \textbf{0.731 ± 0.045}  \\

\bottomrule
\end{tabular}}
\vspace{-10pt}
\end{center}}
\end{table}

As we discussed in the main paper, we propose a novel diffusion feedback-driven loss $\mathcal{L}_{fd}$ to further facilitate the disentanglement of the representations. Formally, we make the original $C$ in $\beta$-VAE auto-changeable, and control its value dynamically according to the information capacity of the reverse diffusion process as follows:

{{\footnotesize
\begin{equation}
    C_{dyn} = f({E_{x_t}}) = \begin{cases}
  C_{base} \cdot (\frac{E_{x_0}}{E_{x_t}}),& \text{ if } 0<f({E_{x_t}})<C_{max} \\
  C_{max}, &\text{ if } f({E_{x_t}}) \geq C_{max}
\end{cases}
\end{equation}}\normalsize
{\footnotesize
\begin{equation}
E_{{x_0}} = -\sum_{ p_{{x_0}} \in \mathbb{R}} p_{{x_0}}logp_{{x_0}}, 
\hspace{2mm}
E_{{x_t}} = -\sum_{p_{{x_t}} \in \mathbb{R}} p_{{x_t}}logp_{{x_t}}, 
\end{equation}}\normalsize}

\noindent where $C_{base}$, $C_{max}$ are the original empirical start-point and the maximum upper-bound according to~\cite{burgess2018understanding}, and their influences are evaluated as follows. The results depicted in Table~\ref{abl_c} demonstrate that our method is robust enough to achieve a stable satisfactory performance with different values of $C_{base}$, $C_{max}$. Besides, these experimental results reveal that setting $C_{base}$ to 10 consistently yields optimal disentanglement outcomes. Moreover, our assessment also extends to examining the impact of the upper-bound value, $C_{max}$. As shown in Table~\ref{abl_c}, reducing the gap between $C_{max}$ and $C_{base}$ hampers the overall disentanglement performance (0.794 vs. 0.762 in FactorVAE score). We analyze that, the variable scope of the dynamic $C_{dyn}$ will be limited due to the relatively small range between $C_{max}$ and $C_{base}$. This limitation may negatively impact the control of KL-divergence, consequently hampering the overall model's ability to effectively disentangle the inner latent space more.


\subsection{More Subjective Results}

{In this section, we further provide more visualization results to demonstrate the generalization ability of our CL-Dis on different datasets}, including FFHQ~\cite{karras2019style}, LSUN Cars~\cite{yu2015lsun} and CelebA~\cite{karras2017progressive}. 
As shown in Figure~\ref{supp_vis}, our CL-Dis could generate fine-grained disentangled results along the \textit{``glasses, mouth, hair, car color, hair color, and face illumination''} attribute changes across the traversal direction, which fully illustrates the well-generalized disentanglement capability of our method.

\subsection{Dimension Ablation on Shape3D}
{We further conduct Dimension Ablation on Shapes3D dataset in Table~\ref{dim}}. We can see below that reducing the dimension of $z_{sem}$ leads to minor changes, which indicates our method is robust enough.

\begin{table}[th]
\footnotesize
\centering
\vspace{-5pt}
\begin{center}
\renewcommand{\arraystretch}{1.0}
\setlength{\tabcolsep}{4.8mm}{
\begin{tabular}{ccccccc}
\toprule
Dimension & FactorVAE score & DCI \\
\midrule
CL-Dis (512)  &   0.840 ± 0.066  &   0.611 ± 0.082     \\
CL-Dis (100) &  0.842 ± 0.042   & 0.547  ± 0.051  \\
CL-Dis (32)  &  0.837 ± 0.053  & 0.582 ± 0.049  \\  
\bottomrule
\end{tabular}}
\vspace{-10pt}
\caption{Ablation study for the dimension of $z_{sem}$ on the Shapes3D dataset.}
\label{dim}
\end{center}
\vspace{-4mm}
\end{table}

\section{Limitation Discussions}

In Table~\ref{disqnti2} (corresponds to Table 2 in the main manuscript), we observe our method encounters an unsatisfactory performance on the MPI3D dataset, which indicates that our method may be distracted by the real-world natural environment scenarios. We discuss the potential reason that the collected data in MPI3D~\cite{gondal2019transfer},  focus on bridging the gap between simulation and reality, which distinguishes it apart from various other synthetic toy datasets~\cite{reed2015deep,kim2018disentangling}. In detail, real-world recordings in MPI3D introduce more extra environment interruptions, such as chromatic aberrations in cameras and complex surface properties of objects (e.g., reflectance, radiance, and irradiance). These interruptions pose challenges in learning disentangled representations due to the mixing of factors and distortions during the real-world recordings, especially for scalar-based methods like ours~\cite{gondal2019transfer}. This unsolved limitation of our method will be left as future research work.
{
    \small
    \bibliographystyle{ieeenat_fullname}
    \bibliography{main}
}


\end{document}